\documentclass{article}
\usepackage{spconf,amsmath,graphicx}
\usepackage{multirow}

\usepackage{amsmath} 
\usepackage[subnum]{cases}

\usepackage[font=small,skip=5pt]{caption}

\title{ ENHANCING AUTOMATICALLY DISCOVERED MULTI-LEVEL ACOUSTIC PATTERNS CONSIDERING CONTEXT CONSISTENCY WITH APPLICATIONS IN SPOKEN TERM DETECTION}

\name{Cheng-Tao Chung$^{\#1}$, Wei-Ning Hsu$^{*2}$, Cheng-Yi Lee$^{*3}$, and Lin-Shan Lee$^{\#4}$}
\address{
Graduate Institute of Electrical Engineering, National Taiwan University$^{\#}$\\
Department of Electrical Engineering, National Taiwan University$^{*}$\\
\small
\texttt{b97901182@gmail.com$^{1}$, mhng1580@gmail.com$^{2}$, chenyi2229@gmail.com$^{3}$, lslee@gate.sinica.edu.tw$^{4}$}
\normalsize
}
\begin{document}
\fontsize{8.9}{10}\selectfont


\setlength{\textfloatsep}{2pt plus 1.0pt minus 1.0pt}
\setlength{\floatsep}{2pt plus 1.0pt minus 1.0pt}
\setlength{\intextsep}{1.2pt plus 1.0pt minus 1.0pt}
\maketitle


\begin{abstract}
This paper presents a novel approach for enhancing the multiple sets of acoustic patterns automatically discovered from a given corpus. In a previous work it was proposed that different HMM configurations (number of states per model, number of distinct models) for the acoustic patterns form a two-dimensional space. Multiple sets of acoustic patterns automatically discovered with the HMM configurations properly located on different points over this two-dimensional space were shown to be complementary to one another, jointly capturing the characteristics of the given corpus. By representing the given corpus as sequences of acoustic patterns on different HMM sets, the pattern indices in these sequences can be relabeled considering the context consistency across the different sequences. Good improvements were observed in preliminary experiments of pattern spoken term detection (STD) performed on both TIMIT and Mandarin Broadcast News with such enhanced patterns.

\end{abstract}

\section{Introduction}
Supervised training of HMMs for large vocabulary continuous speech recognition (LVCSR) relies on not only collecting huge quantities of acoustic data, but also obtaining the corresponding transcriptions. Such supervised training methods yield adequate performance in most circumstances but at high cost, and in many situations such annotated data sets are simply not available.  This is why substantial effort \cite{wang2014graph}\cite{jansen2011towards}\cite{lee2012nonparametric}\cite{gish2009unsupervised}\cite{siu2010improved}\cite{chung2013unsupervised}\cite{chan2013toward} has been made for unsupervised discovery of acoustic patterns from huge quantities of acoustic data without annotation, which may be easily obtained nowadays. For some applications such as Spoken Term Detection (STD) \cite{saraclar2004lattice}\cite{miller2007rapid}\cite{mamou2007vocabulary}\cite{wallace2007phonetic}\cite{pan2010performance} in which the goal is simply to match and find some signal segments, the extra effort of building an LVCSR system using corpora with human annotations is very often an unnecessary burden \cite{metze2012spoken}\cite{jansen2012indexing}\cite{zhang2009unsupervised}\cite{huijbregts2011unsupervised}\cite{wang2012acoustic}. M
ost effort of unsupervised discovery of acoustic patterns considered only one level of phoneme-like acoustic patterns. However, it is well known that speech signals have multi-level structures including at least phonemes and words, and such structures are very helpful in analysing or decoding speech \cite{pan2010performance}.
In a previous work, we proposed to discover the hierarchical structure of two-level acoustic patterns, including subword-like and word-like patterns. A similar two-level framework was also developed recently \cite{walter2013hierarchical}. In a more recent attempt \cite{chung2014unsupervised}, we further proposed a framework of discovering multi-level acoustic patterns with varying model granularity. 
The different pattern HMM configurations (number of states per model, number of distinct models) form a two-dimensional model granularity space. Different sets of acoustic patterns with HMM model configurations represented by different points properly distributed over this two-dimensional space are complementary to one another, thus jointly capture the characteristics of the corpora considered. Such a multi-level framework was shown to be very helpful in the task of unsupervised spoken term detection (STD) with spoken queries, because token matching can be performed with pattern indices on different levels of signal characteristics, and the information integration across multiple model granularities offered the improved performance.

In this work, we further propose an enhanced version of the multi-level acoustic patterns with varying model granularity by considering the context consistency for the decoded pattern sequences within each level and across different levels. In other words, the acoustic patterns discovered on different levels are no longer trained completely independently. We try to ``relabel" the pattern sequence for each utterance in the training corpora considering the context consistency within and across levels. For a certain level, the context consistency may indicate that the realizations of a certain pattern should be split into two different patterns, while the realizations of another two patterns should be merged. In this way the multi-level acoustic patterns can be enhanced.

\section{Proposed Approach}
\subsection{Pattern Discovery for a Given Model Configuration}

Given an unlabeled speech corpus, it is not difficult for unsupervised discovery of the desired acoustic patterns from the corpus for a chosen hyperparameter set $\psi$ that determines the HMM configuration (number of states per model and number of distinct models) \cite{jansen2011towards}\cite{gish2009unsupervised}\cite{siu2010improved}\cite{chung2013unsupervised}\cite{creutz2007unsupervised}.
This can be achieved by first finding an initial label $\omega_0$ based on a set of assumed patterns for all observations in the corpus $\chi$ as in (\ref{eq:1}) \cite{chung2013unsupervised}.
Then in each iteration $t$ the HMM parameter set $\theta^\psi_{t}$ can be trained with the label $\omega_{t-1}$ obtained in the previous iteration as in (\ref{eq:2}), and the new label $\omega_{t}$ can be obtained by pattern decoding with the obtained parameter set $\theta^\psi_{t}$ as in (\ref{eq:3}). 
\begin{eqnarray}
\omega_{0}&=& \mbox{initialization}(\chi),\phantom{\arg \max_{\substack{\theta^\psi}}}                                           \label{eq:1} \\ 
\theta^\psi_{t} &=& \arg \max_{\substack{\theta^\psi}} P(\chi|\theta^\psi,\omega_{t-1}),             \label{eq:2} \\
\omega_{t} &=& \arg \max_{\substack{\omega}} P(\chi|\theta^\psi_{t} ,\omega).                        \label{eq:3}
\end{eqnarray}
The training process can be repeated with enough number of iterations until a converged set of pattern HMMs is obtained.

\subsection{Model Granularity Space for Multi-level Pattern Sets}

\begin{figure}[t]
\centerline{\includegraphics[width=0.4\textwidth]{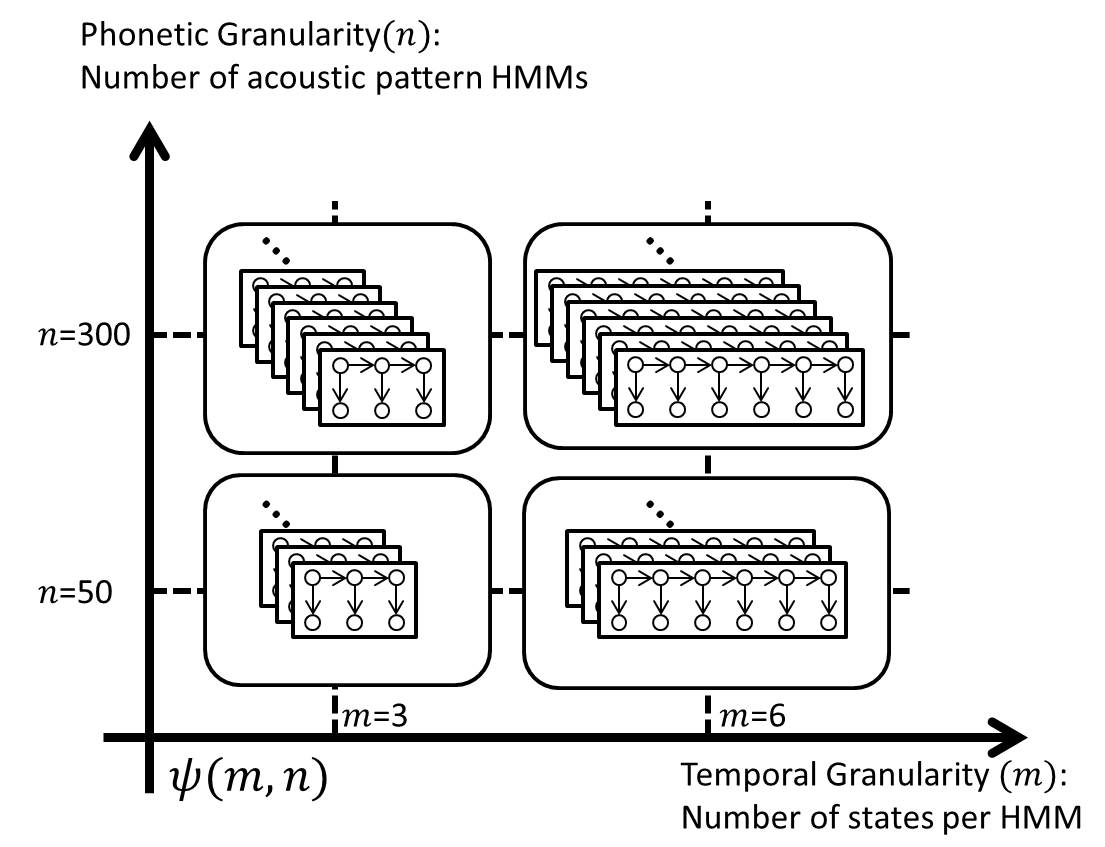}}
\caption{Model granularity space for acoustic pattern configurations}\label{fig:2dcube}
\end{figure}

The above process can be performed with many different HMM configurations, each characterized by two hyperparameters: the number of states $m$ in each acoustic pattern HMM, and the total number of distinct acoustic patterns $n$ during initialization,  $\psi=(m,n)$. The transcription of a signal decoded with these patterns can be considered as a temporal segmentation of the signal, so the HMM length (or number of states in each HMM) $m$ represents the temporal granularity. The set of all distinct acoustic patterns can be considered as a segmentation of the phonetic space, so the total number $n$ of distinct acoustic patterns represents the phonetic granularity. 
This gives a two-dimensional representation of the acoustic pattern configurations in terms of temporal and phonetic granularities as in Fig. \ref{fig:2dcube}. Any point in this two-dimensional space in Fig. \ref{fig:2dcube} corresponds to an acoustic pattern configuration. 
Note that in our previous work \cite{chung2014unsupervised}, the effect of the third dimension, the acoustic granularity which is the number of Gaussians in each state, was shown to be negligible, thus here we simply set the number of Gaussians in each state to be 4 in all cases. Although the selection of the hyperparameters can be arbitrary in this two-dimensional space, here we only select $M$ temporal granularities and $N$ phonetic granularities, forming a two-dimensional array of $M \times N$ hyperparameter sets in the granularity space.

\subsection{Pattern Relabeling Considering Context Consistency}

\begin{figure}[b]
\centerline{\includegraphics[width=0.4\textwidth]{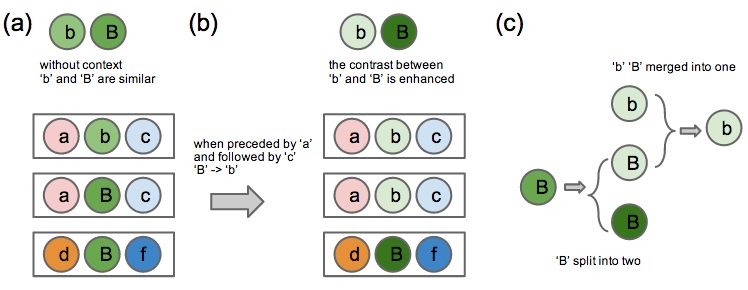}}
\caption{Pattern relabeling considering context consistency}\label{fig:exp}
\end{figure}

Context constraints successfully explored in language modeling can be used here for relabeling the acoustic patterns as shown by an example in Fig. \ref{fig:exp}. We assume the patterns `b' and `B' are similar without context as in Fig. \ref{fig:exp}(a). However if the context is considered, we may observe from the corpus that many realizations of pattern `b' is preceded by pattern `a' and followed by pattern `c', while most realizations of pattern `B' have different context. Therefore by relabeling all realizations of pattern `B' which are preceded by pattern `a' and followed by pattern `c' as pattern `b', the contrast between patterns `b' and `B' can be enhanced during the next iteration of acoustic model update as shown in Fig. \ref{fig:exp}(b) since the borderline cases have been resolved. As shown in Fig. \ref{fig:exp}(c), this relabeling includes both pattern splitting and merging, since the realizations of pattern `B' are split into two patterns `B' and `b', while some realizations of pattern `B' are merged into pattern `b'. The example here considers the context in time, but can be generalized to context in model granularities as explained below.

\begin{figure}[t]
\centerline{\includegraphics[width=0.4\textwidth]{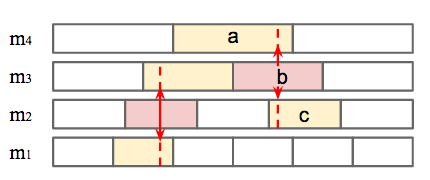}}
\caption{Local smoothing considering granularity context}\label{fig:tile}
\end{figure}

As shown in Fig. \ref{fig:tile}, assuming an utterance is decoded into four different pattern sequences using four sets of patterns with neighboring temporal granularity $m_{4}>m_{3}>m_{2}>m_{1}$, i.e., pattern HMMs with different lengths. Considering a realization of pattern `b' of temporal granularity $m_{3}$, we find its central frame belongs to the realization  of pattern `a' of temporal granularity $m_{4}$ and the realization of pattern `c' of temporal granularity $m_{2}$. So patterns `a' and `c' are taken as the context of pattern `b' in neighboring temporal granularities. The same could be done for phonetic granularity.

\subsection{Pattern Relabeling Method}
Let $\omega(m_{k},n_{k},l)$ be the index for a decoded acoustic pattern at time $l$ within an utterance in the corpus $\chi$ using the acoustic pattern set with the granularity $\psi(m_{k},n_{k})$. The relabeled pattern $\overline{\omega}(m_{k},n_{k},l)$ is then as in (\ref{eq:4}) i.e., the pattern among all patterns in the set of $\psi(m_{k},n_{k})$ which maximizes the product of the three probabilities in (\ref{eq:5-1})(\ref{eq:5-2})(\ref{eq:5-3}) evaluated with the context respectively in $l$, $n$, and $m$. The first probability $P_{l}(w)$ in (\ref{eq:5-1}) for context in time $l$ is actually the product of forward bigram and backward bigram well known in language modeling. The other two probabilities $P_{n}(w)$, $P_{m}(w)$ in (\ref{eq:5-2})(\ref{eq:5-3}) are exactly the same, except $n_{k-1}$, $n_{k+1}$ and $m_{k-1}$, $m_{k+1}$ are the neighboring values of $n$ and $m$.
\begin{subequations}
\begin{align}
\overline{\omega}(m_{k},n_{k},l) =\arg \max_{w}(P_{l}(w)  P_{n}(w) P_{m}(w)),	\phantom{ \max_{w}} \label{eq:4}\\
P_{l}(w) = P(w|\omega(m_k,n_k,l\mbox{-1}))  P(w|\omega(m_k,n_k,l\mbox{+1})),	\phantom{\max_{w}}	\label{eq:5-1}\\
P_{n}(w) = P(w|\omega(m_k,n_{k-1},l))    P(w|\omega(m_k,n_{k+1},l)),	\phantom{ \max_{w}}	\label{eq:5-2}\\
P_{m}(w) =P(w|\omega(m_{k-1},n_k,l))    P(w|\omega(m_{k+1},n_{k},l)).\phantom{ \max_{w}}		\label{eq:5-3}
\end{align}\label{eq:z}
\end{subequations}
Finer patterns and coarser patterns are drastically different in terms of perplexity; shorter patterns and longer patterns produce very different pattern sequences in terms of duration. They are complementary to each other, but we only consider the context consistency among the neighboring granularity configurations as in \eqref{eq:z}. This relabeling is performed on every decoded sequence of the $M \times N$ pattern sets considered. Katz smoothing \cite{chen1996empirical} was applied to deal with unseen pattern bigrams. On the boundary of the granularity configurations or time sequences, the bigram probability is taken as 1. 

\subsection{Pattern Enhancement by Re-estimation after Relabeling}
The relabeling in (\ref{eq:4}) can be inserted into the recursive process of discovering the patterns in each iteration in (\ref{eq:2})(\ref{eq:3}), as shown in (\ref{eq:7})(\ref{eq:8}).
\begin{eqnarray}
\overline{\omega}_{t} &=& \arg \max_{\substack{\overline{\omega}}} P(\overline{\omega}|\omega_{t}),             \label{eq:7} \\
\theta^\psi_{t+1} &=& \arg \max_{\substack{\theta^\psi}} P(\chi|\theta^\psi,\overline{\omega}_{t}).             \label{eq:8} 
\end{eqnarray}
When an iteration is completed as in (\ref{eq:2})(\ref{eq:3}), a new set of patterns is generated as in (\ref{eq:2}), with which a new set of labels is obtained as in (\ref{eq:3}).  The new labels $\omega_{t}$ in (\ref{eq:3}) is then relabeled with (\ref{eq:4}) based on the new labels $\omega_{t}$ on all different HMM sets to produce a slightly better label $\overline{\omega_{t}}$ as in (\ref{eq:7}). This slightly better label $\overline{\omega_{t}}$ is then used in (\ref{eq:8}) to generate a slightly better model set $\theta^\psi_{t+1}$. Note that (\ref{eq:8}) is almost the same as (\ref{eq:2}), except here based on the slightly better label $\overline{\omega_{t}}$ obtained in (\ref{eq:7}). In this way the relabeling process can be repeatedly applied in every iteration, and the patterns can be enhanced by the relabeling process during the model re-estimation. Although it is theoretically possible to consider the optimization process in (\ref{eq:3}) and (\ref{eq:7}) jointly in a single step, such as maximizing the product of the two probabilities in the right hand sides of (\ref{eq:3}) and (\ref{eq:7}), practically such a joint optimization is computationally unfeasible. Therefore this is done in two separate steps here.


\subsection{Spoken Term Detection}

There can be various applications for the acoustic patterns presented here. In this section we summarize the way to perform spoken term detection \cite{chung2014unsupervised}.
Let \{$p_r, r=1,2,3,..,n$\} denote the $n$ acoustic patterns in the set of $\psi$=$(m,n)$. We first construct a similarity matrix $S$ of size $n \times n$ off-line for every pattern set $\psi$=$(m,n)$, for which the element $S(i,j)$ is the similarity between any two pattern HMMs $p_i$ and $p_j$ in the set.
\begin{equation}
S(i, j) =\mbox{exp}(-\mbox{KL}(i, j)/\beta. \label{eq:soft}
\end{equation}
The KL-divergence $\mbox{KL}(i,j)$ between two pattern HMMs in (\ref{eq:soft}) is defined as the symmetric KL-divergence between the states based on the variational approximation \cite{hershey2007approximating} summed over the states. To transform the KL divergence into a similarity measure between 0 and 1, a negative exponential was applied \cite{marszalek2008constructing} with a scaling factor $\beta$. When $\beta$ is small, similarity between distinct patterns in (\ref{eq:soft}) approaches zero, so (\ref{eq:soft}) approaches the delta function $\delta(i,j)$. $\beta$ can be determined with a held out data set, but here we simply set it to 100.

\begin{figure}[b]
\centerline{\includegraphics[width=0.4\textwidth]{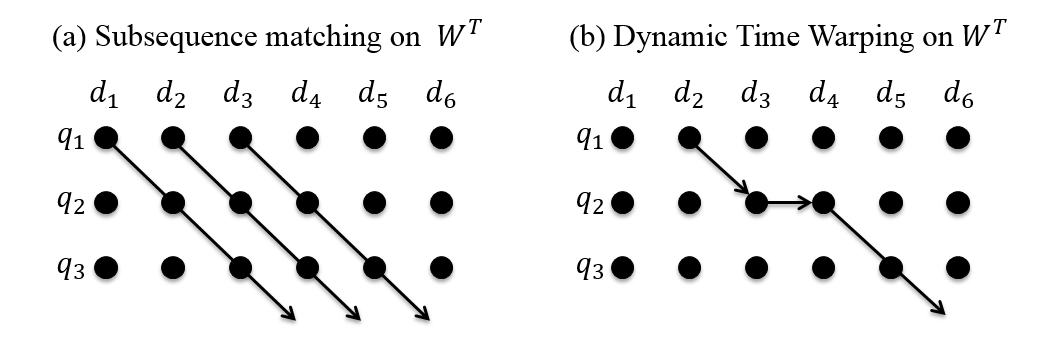}}
\caption{The matching matrix $W$}\label{fig:quant}
\end{figure}

In the on-line phase, we perform the following for each entered spoken query $q$ and each document (utterance) $d$ in the archive for each pattern set $\psi$=$(m,n)$. Assume for a given pattern set a document $d$ is decoded into a sequence of $D$ acoustic patterns with indices $(d_1, d_2, ..., d_D)$ and the query $q$ into a sequence of $Q$ patterns with indices $(q_1, ..., q_Q)$. 
We thus construct a matching matrix $W$ of size $D \times Q$ for every document-query pair, in which each entry $(i,j)$ is the similarity between acoustic patterns with indices $d_i$ and $q_j$ as in (\ref{eq:topk}) and shown in Fig. \ref{fig:quant}(a) for a simple example of $Q=3$ and $D=6$, where $S(i,j)$ is defined in (\ref{eq:soft}),
\begin{equation}
W(i, j)  = S(d_i, q_j).  \label{eq:topk}
\end{equation}
It is possible to consider the N-best pattern sequences rather than the one-best sequences here by considering the posteriorgram vectors based on the N-best sequences for $d$, $q$ and integrate them in the matrix $W$. However, previous experiments showed that the extra improvements brought in this way is almost negligible, probably because the $M \times N$ different pattern sequences based on the $M \times N$ different pattern sets can be considered as a huge lattice including many one-best paths which will be jointly considered here \cite{chung2014unsupervised}.

For matching the sub-sequence of $d$ with $q$, we sum the elements in the matrix $W$ in \eqref{eq:dtw}  along the diagonal direction, generating the accumulated similarities for all sub-sequences starting at all pattern positions in $d$ as shown in Fig. \ref{fig:quant}(a). The maximum is selected to represent the relevance between document $d$ and query $q$ on the pattern set $\psi$=$(m,n)$ as in (\ref{eq:dtw}).
\begin{equation}
R(d,q) = 
\max_{\substack{i }}\sum_{j=1}^{Q} W(i+j,j).  \label{eq:dtw}
\end{equation}
It is also possible to consider dynamic time warping (DTW) on the matrix $W$ as shown in Fig. \ref{fig:quant}(b). However, previous experiments showed that the extra improvements brought in this way is almost negligible, probably because here we have jointly considered the $M \times N$ different pattern sequences based on the $M \times N$ different pattern sets (e.g. including longer /shorter patterns), so the different time-warped matching and insertion/deletion between $d$ and $q$ is already automatically included \cite{chung2014unsupervised}.

The $M \times N$ relevance scores $R(d,q)$ in (\ref{eq:dtw}) obtained with $M \times N$ pattern sets $\psi$=$(m,n)$ are then averaged and the average scores are used in ranking all the documents for spoken term detection. It is also possible to learn the weights for different pattern sets to produce better results using a development set. But here we simply assume the detection is completely unsupervised without any annotation, and all pattern sets are equally weighted \cite{chung2014unsupervised}.

\section{Experiments}

\subsection{Purity in Pattern Sequences for known Words}
\begin{figure}[t]
\centerline{\includegraphics[width=0.4\textwidth]{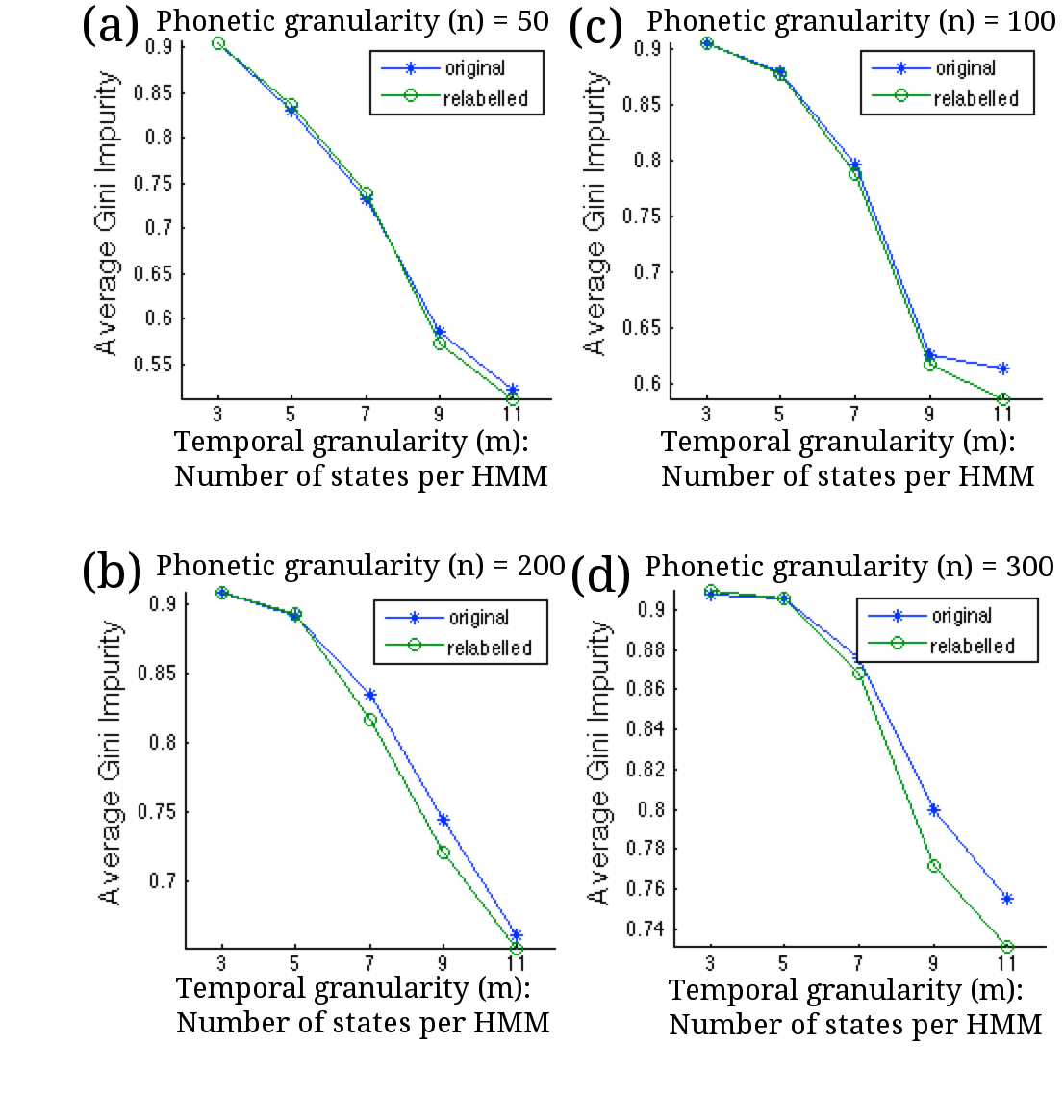}}
\caption{Average Gini impurity for the top 20 words with the highest counts in the TIMIT training set based on the original patterns (blue) and those after relabeling (green), with different values of $n$ and $m$.}\label{fig:t_s_20}
\end{figure}
%
%
%
%
In order to evaluate the quality of the acoustic patterns we discovered with varying temporal and phonetic granularities, we use the Gini impurity for the pattern sequences found for known high frequency words, since this can be evaluated for any given pattern set. Assume all the realizations of a high frequency word (e.g. the word ``water") are decoded into $I$ different pattern sequences, each occupying a percentage $f_i$ of the realizations ($\Sigma_i f_i =1$), we can evaluate the Gini impurity \cite{breiman1996technical} for the word using the $I$ percentages $f$=$\{f_i,i\mbox{=1,2,...$I$}\}$ as in (\ref{eq:9}), 
\begin{equation}
\mbox{Gini Impurity}(f) = \sum_{i=1}^{I}{f_{i}(1-f_{i})}.		\label{eq:9}
\end{equation}
Gini impurity falls within the interval $[0,1)$, reaches zero when all the realizations are decoded into the same pattern sequence, and becomes larger when the distribution is less pure.
We trained the above different sets of patterns with $m$=3, 5, 7, 9, 11 and $n$=50, 100, 200, 300 on the TIMIT training set. Fig. \ref{fig:t_s_20} shows the average Gini impurity for the top 20 words with the highest occurrence counts in TIMIT training set, based on the original patterns (blue) and those after relabeling (green) for all cases considered. 
We see the impurity was in general high for such automatically discovered patterns because the realizations of the same phoneme produced different speakers were possibly decoded as different patterns,  and the insertion/deletion inevitably increased the impurity. 
Although the impurity was high, the relabeling proposed here generated better patterns. We see the difference was more significant for larger $m$. 
Because the temporal variation is easily captured by models with short patterns ($m$=3 or 5 with high impurity) which increases the impurity, much lower impurity was achieved with longer patterns ($m$=9 or 11).

\begin{figure}[t]
\centerline{\includegraphics[width=0.45\textwidth]{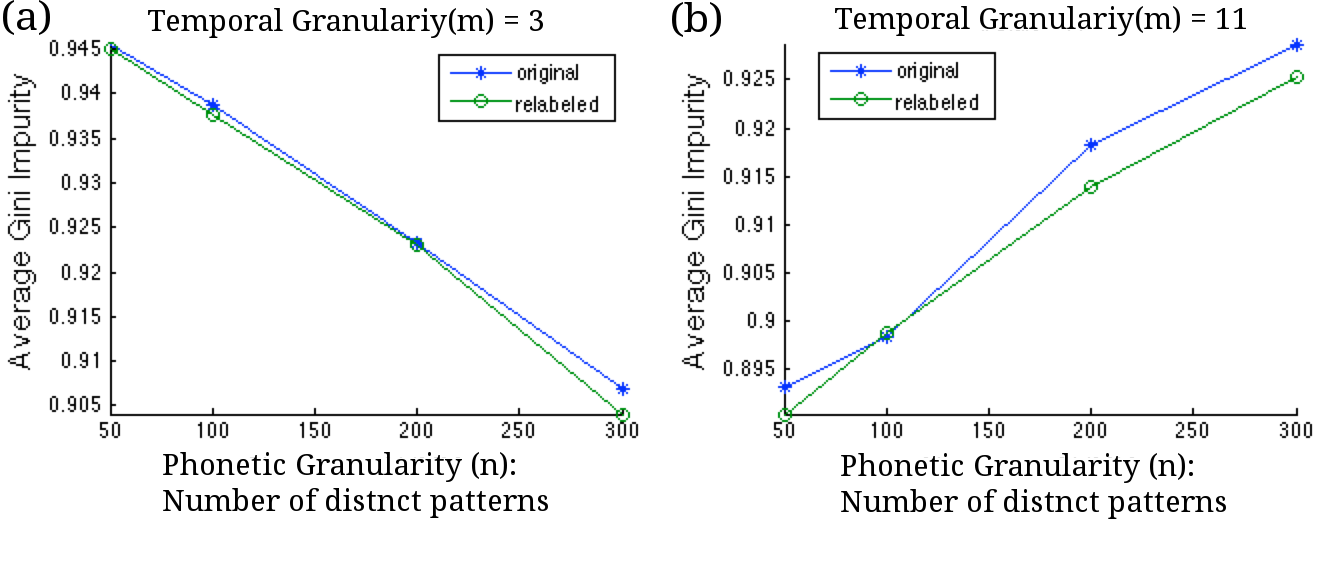}}
\caption{Average Gini impurity for the cluster of words with occurrence counts ranging from 16 to 22 with (a) $m$=3, (b) $m$=11.}\label{fig:c_h_16_23}
\end{figure}

Another set of results for average Gini impurity for the cluster of words with occurrence counts ranging from 16 to 22 in the TIMIT training set is shown in Fig. \ref{fig:c_h_16_23} for $m$=3 and 11 states per HMM with varying number of distinct patterns $(n)$. It is still quite clear that the relabeling process enhanced the patterns, and it is interesting to note that the trends for $m$=3 and 11 are quite different (Fig. \ref{fig:t_s_20}(a) and (b)). As mentioned above, the temporal variation is easily captured by models with short patterns which increases the impurity
 (e.g. $m$=3 in Fig. \ref{fig:c_h_16_23}(a)) so increasing the number of patterns $(n)$ helped reduce the impurity. However, when the models are long enough (e.g. $m$=11 in Fig. \ref{fig:c_h_16_23}(b)), larger number of patterns($n$) gives more redundant patterns which caused confusion during decoding, so the impurity went up with larger $n$. These results indicate that the different sets of patterns of different model granularities were complementary to each other. 
Note that only high frequency words with enough realizations can be used or the impurity evaluation here to show the quality of the patterns. But how these patterns can be applied to spoken term detection will be shown below, for which the queries are usually low frequency words, whose impurity is difficult to evaluate.



%
%

\subsection{Unsupervised Spoken Term Detection}
We conducted two separate query by example spoken term detection experiments on two spoken archives. In the first experiment, the TIMIT training set was used as the spoken archive and the spoken query set consisted of 16 words randomly selected from the TIMIT testing set. In the second experiment, the spoken archive was 4.5 hours of Mandarin Broadcast News segmented into 5034 spoken documents and the spoken query set was 10 words selected from another development set. In either case, a spoken instance of a query word was randomly selected from the data set, and used as the spoken query to search for other instances in the spoken archive.  The conventional 39 dimensional MFCC features were used for the HMMs. 20 sets of acoustic patterns were generated for TIMIT with $m$ = 3, 5, 7, 9, 11 and $n$ = 50, 100, 200, 300; 9 sets for the Mandarin Broadcast News with $m$ = 3, 7, 13 and $n$ = 50, 100, 300; all with 4 Gaussian mixtures per state. 
We compared $\overline{\omega}(m,n)$ with $\omega(m,n)$ for each $(m,n)$ pair. 
 We used the mean average precision (MAP) \cite{philbin2007object}\cite{yue2007support} as the performance measure, a higher value implies better performance.

\begin{figure}[b]
\centerline{\includegraphics[width=0.50\textwidth]{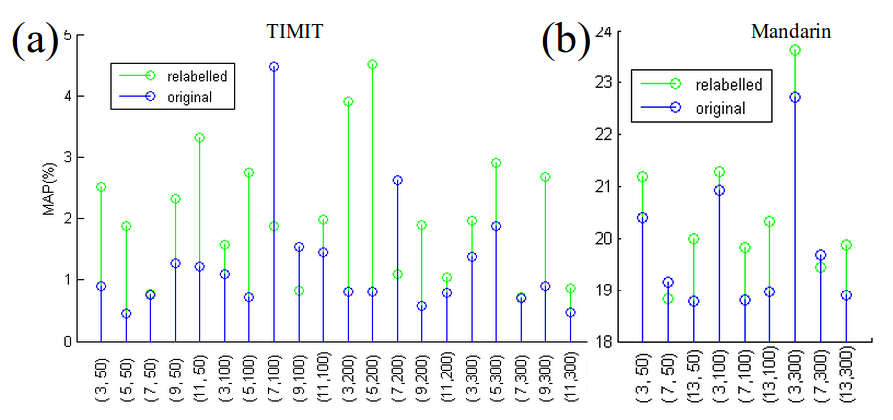}}

\caption{mean average precision of each of the HMM sets with the granularity hyperparameters ($m$,$n$) on (a) TIMIT and (b) Mandarin Broadcast News.}\label{fig:26}
\end{figure}

The MAP performance of each of the 20 pattern sets for TIMIT and 9 sets for Mandarin Broadcast News before and after relabeling is in Fig. \ref{fig:26}(a)(b) where the performance was clearly boosted for most of the pattern sets. A paired sample t-test was used to check the MAP improvement of relabeled pattern sets, $t$(28)=3.37, $p$=0.0011, significant improvement was observed.
Note that different from TIMIT which had many different speakers, the Mandarin Broadcast News was produced by a limited number of anchors, so MAP for each pattern set ranged between 18\% to 22\%, much higher than TIMIT.
 Although the MAP for each individual pattern set was relatively low on TIMIT (1\% to 5\%) in general, much better results in MAP can be obtained when all of them are jointly  considered as rows (b)(c) in Table 1.  
Row (a) in Table \ref{tab:1} was the frame-based dynamic time warping (DTW) on MFCC sequences. We see the relabeled patterns achieved an MAP of 28.26\% and  24.50\% which is significantly better than that using the original patterns (26.32\% and 23.38\%). 
 Further more, both of them significantly outperformed the baseline (10.16\% and 22.19\%), which proved the improvement was non-trivial.

\begin {table}[t]
\begin{center}
    \begin{tabular}{ | l | p{1.2cm} | p{1.2cm}|}
    \hline
    Search methods(MAP) & TIMIT & Mandarin \\ \hline
    (a) frame-based DTW on MFCC & 10.16\% & 22.19\% \\ \hline
    (b) proposed: original patterns & 26.32\% & 23.38\% \\ \hline
    (c) proposed: relabeled patterns &  28.26\% & 24.50\% \\
    \hline
    \end{tabular}
\end{center}
\caption {Overall spoken term detection performance in mean average precision.\label{tab:1}}
\end{table}

\section{Conclusion}
In this work, we propose a method for improving the quality of multi-level acoustic patterns discovered from a target corpus. By incorporating context consistency in time and model granularity, a more consistent set of patterns can be obtained. This is verified with improved performance in spoken term detection on TIMIT and Mandarin Broadcast News.

{
\fontsize{9}{9}\selectfont
\bibliography{strings,refs}
\include{refs}

\bibliographystyle{IEEEbib}

}

\end{document}